\documentclass[11pt]{article}

\usepackage[preprint]{acl}

\usepackage{times}
\usepackage{latexsym}

\usepackage[T1]{fontenc}

\usepackage[utf8]{inputenc}

\usepackage{microtype}

\usepackage{inconsolata}

\usepackage{graphicx}

\usepackage{amsmath}
\usepackage{amssymb}

\usepackage{booktabs}
\usepackage{wrapfig}
\usepackage{float}
\usepackage[most]{tcolorbox}
\usepackage[table]{xcolor}

\title{Lizard: An Efficient Linearization Framework for Large Language Models}

\author{
  Chien Van Nguyen$^{1}$ \quad
  Huy Nguyen$^{1}$ \quad
  Ruiyi Zhang$^{2}$ \quad
  Hanieh Deilamsalehy$^{2}$ \\
  \textbf{Puneet Mathur$^{2}$} \quad
  \textbf{Viet Lai$^{2}$} \quad
  \textbf{Haoliang Wang$^{2}$} \quad
  \textbf{Jayakumar Subramanian$^{2}$} \\
  \textbf{Ryan A. Rossi$^{2}$} \quad
  \textbf{Trung Bui$^{2}$} \quad
  \textbf{Nikos Vlassis$^{2}$} \quad
  \textbf{Franck Dernoncourt$^{2}$\thanks{Corresponding authors}} \quad
  \textbf{Thien Huu Nguyen$^{1}$\footnotemark[1]}
  \\[2ex] %
  $^1$University of Oregon \quad 
  $^2$Adobe Research \\[2ex]
  \texttt{\{chienn,thienn\}@uoregon.edu} \quad
  \texttt{\{ryrossi,bui,vlassis,dernonco\}@adobe.com}
}

\begin{document}
\maketitle
\begin{abstract}
We propose Lizard, a linearization framework that transforms pretrained Transformer-based Large Language Models (LLMs) into subquadratic architectures. 
Transformers faces severe computational and memory bottlenecks with long sequences due to the quadratic complexity of softmax attention and the growing Key-Value (KV) cache that makes inference memory-bound by context length. 
Lizard addresses these limitations by introducing a subquadratic attention mechanism that closely approximates softmax attention while preserving model quality. 
Unlike prior linearization methods constrained by fixed, non-adaptive structures, Lizard augments the architecture with compact, learnable modules that enable adaptive memory control and robust length generalization. Moreover, we introduce a hardware-aware algorithm that solves numerical instability in gated attention to accelerate training. Extensive experiments show that Lizard achieves near-lossless recovery of its teacher model's performance, significantly outperforming previous methods by up to 9.4 - 24.5 points on the 5-shot MMLU benchmark and demonstrating superior associative recall.

\end{abstract}

\section{Introduction}

\begin{figure*}[t]
  \centering
  \includegraphics[width=\textwidth]{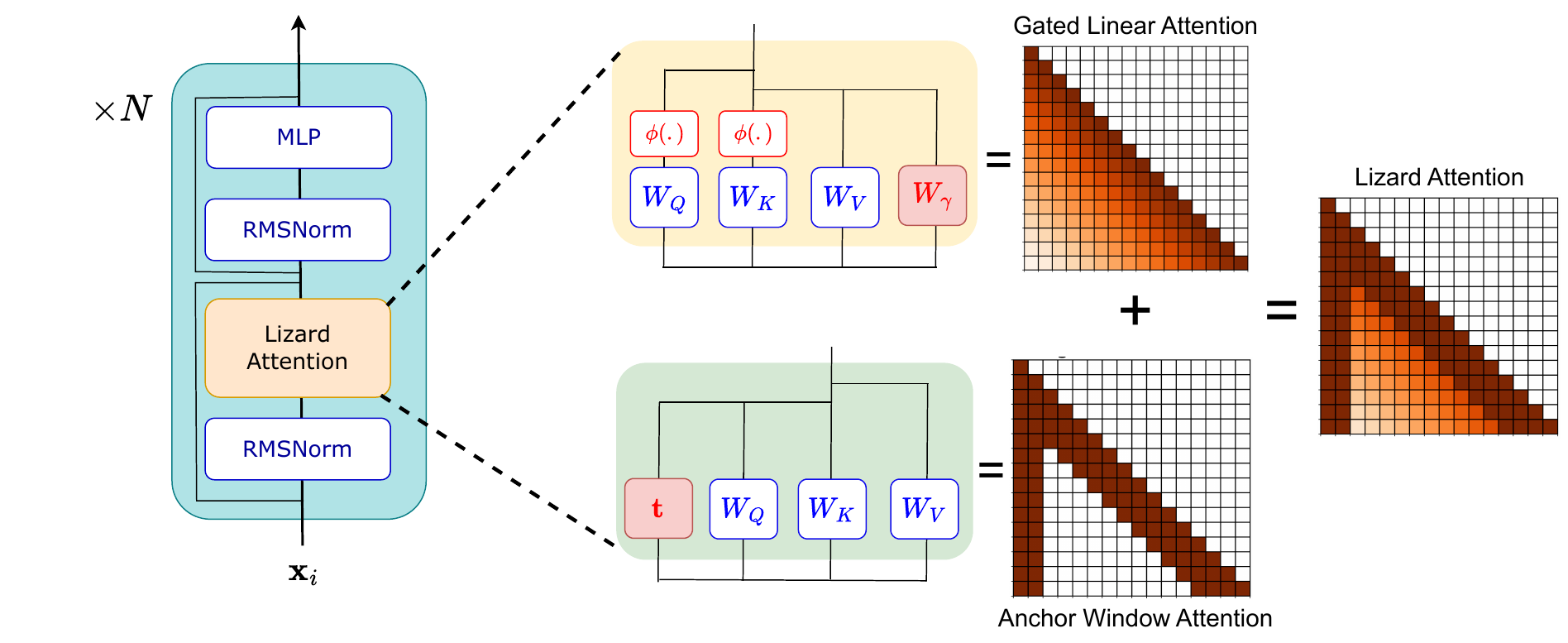}
\caption{An overview of the Lizard Attention architecture. Lizard replaces standard attention with a hybrid mechanism that combines Gated Linear Attention (top) for global context compression and Anchor Window Attention (bottom) for local precision. The components highlighted in red the feature maps ($\phi$), the gating module ($W_\gamma$), and the meta-memory tokens ($t$) represent the compact, learnable modules that are augmented to the teacher architecture.}
  \label{fig:pipeline}
\end{figure*}

Large Language Models (LLMs) based on Transformer architectures have achieved impressive advances in natural language processing across various tasks \citep{grattafiori2024llama, mann2020language, achiam2023gpt}. However, their reliance on softmax attention with quadratic time and memory complexity poses significant limitations for long-context applications, both during training and inference. In particular, the softmax attention computation scales quadratically with the sequence length, while the key-value (KV) cache grows linearly during generation, resulting in significant computational and memory overhead for long-context sequences \citep{adnan2024keyformer}.

Recent work has proposed linear and subquadratic alternatives to softmax attention \citep{yang2024gated, gu2023mamba, peng2023rwkv, wang2020linformer, dao2024transformers}, enabling linear-time training and constant-memory inference.
However, despite these efficiency benefits, pretraining LLMs with such architectures from scratch requires massive training budgets, often involving trillions of tokens.
More importantly, models trained with linear attention mechanisms from scratch consistently underperform on tasks that require in-context learning and retrieval.
For example, Transformer models significantly outperform both Mamba and Mamba-2 by up to 15 points on the 5-shot MMLU benchmark when all are pretrained under the same settings \citep{waleffe2406empirical}. 
Another promising direction involves \emph{linearizing} pretrained Transformer-based LLMs by replacing their softmax attention modules with subquadratic alternatives \citep{zhang2402hedgehog, mercat2024linearizing, zhang2025lolcats, wang2024mamba, lan2025liger}. This strategy aims to retain the rich capabilities of models trained on trillions of tokens while inheriting the efficiency of subquadratic architectures. However, existing linearization methods have consistently fallen short on two key goals: {\bf (1)} preserving performance parity with the teacher model, and {\bf (2)} enabling robust long-context generalization.

These shortcomings stem from two core architectural limitations. {\bf First}, existing approaches lack mechanisms for adaptive memory control. Methods like LoLCATs \citep{zhang2025lolcats} overlook the sophisticated gating mechanisms critical for modulating contextual information in modern recurrent models. Others like Liger \citep{lan2025liger} attempts to a gate but constrains it to a fixed, parameter-free pooling operation. While parameter-efficient, this non-learnable transformation creates an information bottleneck that prevents the model from learning optimal memory dynamics. These architectural choices lead to substantial performance degradation: on the 5-shot MMLU benchmark, LoLCATs lags behind its teacher model by 13.8 points, while Liger exhibits a 21.9 point drop. {\bf Second}, these methods fail at length extrapolation due to their reliance on fixed positional encodings. By retaining Rotary Positional Embeddings (RoPE) \citep{su2024roformer} from the pretrained teacher, these models are restricted to the sequence lengths seen during training. This design fails to leverage the extrapolation capabilities inherent in recurrent formulations and prevents the models from achieving true long-context generation.

In this paper, we introduce \textbf{Lizard} (\underline{\textbf{Li}}neari\underline{\textbf{z}}ing Softmax \underline{\textbf{A}}ttention with \underline{\textbf{R}}ecurrent Gate \underline{\textbf{D}}ynamics), an efficient framework for linearizing LLMs. Unlike prior methods that strictly preserve the teacher's architecture, Lizard adopts a fundamentally different design philosophy: it introduces compact, learnable modules to enable adaptive memory control and data-driven positional modeling, thereby bridging the expressivity-efficiency gap.
At the core of this approach, Lizard is augmented with a learnable gating module that forms a data-adaptive recurrent structure. This module serves two primary purposes. First, it acts as a data-dependent alternative to fixed positional encodings like RoPE, allowing the model to learn relative positional information through adaptive decay patterns for enhanced length generalization. Second, its gated recurrent structure provides a dynamic memory management mechanism, enabling the model to control the retention and forgetting of past tokens to improve associative recall.
Furthermore, to fully recover the expressiveness of softmax attention, the globally-aware gated attention is then combined with Anchor Window Attention, a mechanism that augments local attention with learnable meta-memory tokens. This hybrid framework assigns a specialized role to each component: the gated module captures global context in a compressed form, while the Anchor Window Attention preserves the fine-grained precision of local interactions, resulting in a high-quality approximation of softmax attention.

To complement these architectural innovations, we address a critical implementation bottleneck that hinders the efficiency of gated recurrent models. Standard gated linear attention is often numerically unstable in low-precision formats. This instability forces a reliance on inefficient, full-precision computations, preventing the full utilization of modern hardware accelerators like tensor cores \citep{yang2024gated}. We introduce a hardware-aware algorithm that solves this by reparameterizing the attention computation, making it compatible with tensor core units and improving training throughput by up to 32\%. 

Our contributions are as follows:
\begin{itemize}
    \item We propose Lizard, a linearization framework that converts pretrained Transformers into subquadratic architectures with near-lossless performance recovery. Lizard significantly outperforms prior methods, improving by up to 9.4 - 24.5 points on 5-shot MMLU and demonstrating superior associative recall on long-context retrieval tasks. Furthermore, we show that in a hybrid setup that retains 50\% of the original softmax attention layers, Lizard nearly matches the teacher model's performance on 5-shot MMLU (65.1 vs 66.6)
    \item We introduce a hardware-aware algorithm that solves numerical instability in gated attention, improving training throughput by up to 32\% and enabling more efficient model optimization.
    \item We conduct extensive empirical studies to analyze architectural design choices across diverse benchmarks.
\end{itemize}

\section{Preliminary}
To motivate Lizard, we first review the core components of Causal Softmax Attention, and techniques for linearzing softmax attention.

\noindent \textbf{Causal Softmax Attention: }
In modern Transformer architectures \citep{touvron2023llama, jiang263830494mistral}, for a query vector at position \(i\), Causal Softmax Attention produces the output $\mathbf{y}_i$ as:

\begin{equation*}
    \mathbf{y}_i = \sum_{t=1}^{i}
\frac{
\exp\left(\mathbf{q}_i^\top \mathbf{k}_t / \sqrt{d}\right)
}{
\sum_{j=1}^{i}
\exp\left(\mathbf{q}_i^\top \mathbf{k}_j / \sqrt{d}\right)
} \mathbf{v}_t
\end{equation*}
The expressiveness of this mechanism comes from the softmax function's ability to create a sharp, spiky distribution over past tokens. However, its need to compare every query to all preceding keys results in a computational complexity of $\mathcal{O}(L^2 d)$ for a sequence of length $L$, which is prohibitive for long contexts.

\noindent \textbf{Linear Attention and Linearization} The core idea behind linearization is to replace the expensive softmax function with an efficient alternative. Linear attention mechanisms \citep{katharopoulos2020transformers} achieve this by substituting the exponential similarity function with a kernel function $k(q, k) = \phi(q) \phi(k)^\top$, where $\phi$ is a feature map. The output is then computed as: 

\begin{equation*}
\hat{\mathbf{y}}_i = \frac{\phi(\mathbf{q}_i)^\top \left( \sum_{t=1}^{i} \phi(\mathbf{k}_t) \mathbf{v}_t^\top \right)}{\phi(\mathbf{q}_i)^\top \left( \sum_{j=1}^{i} \phi(\mathbf{k}_j) \right)}
\end{equation*}

This kernel formulation lies at the core of linearized attention, as it enables the reordering of matrix multiplications. Rather than constructing a large $L \times L$ attention matrix, the computation can be reformulated as an incremental update, reducing the complexity to $\mathcal{O}(L d^2)$ and allowing constant-memory inference in a recurrent form:
\[
\mathbf{h}_i = \mathbf{h}_{i-1} + \phi(\mathbf{k}_i) \mathbf{v}_i^\top,\quad \mathbf{y}_i = \phi(\mathbf{q}_i)^\top \mathbf{h}_i
\]

\section{Lizard Framework}

\begin{table*}[t]
  \centering
  \small
   \renewcommand{\arraystretch}{1.0}
  \setlength{\tabcolsep}{0.8pt}
    
  \begin{tabular}{lccccccccc}
    \toprule
    \textbf{Model} & \textbf{Training} & \textbf{PiQA} & \textbf{ARC-e} & \textbf{ARC-c} & \textbf{Hella.} & \textbf{Wino.} & \textbf{MMLU} & \textbf{Avg.} & \textbf{Avg.} \\
    
    & \textbf{Tokens (B)} & acc & acc & acc\_norm & acc\_norm & acc & (5-shot) & (no MMLU) & \\
    
    \midrule
    \multicolumn{10}{l}{\textit{Transformer}} \\
    Gemma-7B & 6000 & 81.9 & 81.1 & 53.2 & 80.7 & 73.7 & 62.9 & 74.1 & 72.3 \\
    Mistral-7B & 8000* & 82.1 & 80.9 & 53.8 & 81.0 & 74.0 & 62.4 & 74.4 & 72.4 \\
    LLaMA-3-8B & 15000 & 79.9 & 80.1 & 53.3 & 79.1 & 73.1 & 66.6 & 73.1 & 72.0 \\
    \midrule
    \multicolumn{10}{l}{\textit{Subquadratic}} \\
    Mamba-7B & 1200 & 81.0 & 77.5 & 46.7 & 77.9 & 71.8 & 33.3 & 71.0 & 64.7 \\
    RWKV-6-v2.1-7B & 1420 & 78.7 & 76.8 & 46.3 & 75.1 & 70.0 & --   & 69.4 & 69.4 \\
    TransNormerLLM-7B & 1400 & 80.1 & 75.4 & 44.4 & 75.2 & 66.1 & 43.1 & 68.2 & 64.1 \\
    Hawk-7B & 300 & 80.0 & 74.4 & 45.9 & 77.6 & 69.9 & 35.0 & 69.6 & 63.8 \\
    Griffin-7B & 300 & 81.0 & 75.4 & 47.9 & 78.6 & 72.6 & 39.3 & 71.1 & 65.8 \\
    \midrule
    \multicolumn{10}{l}{\textit{Linearized (Bounded)}} \\
    Mistral-7B-SUPRA & 100 & 80.4 & 75.9 & 45.8 & 77.1 & 70.3 & 34.2 & 69.9 & 64.0 \\
    Mistral-7B-LoLCATs & 0.04 & 81.5 & 81.7 & 54.9 & 80.7 & 74.0 & 51.4 & 74.5 & 70.7 \\
    LLaMA-3-8B-LoLCATs & 0.04 & 80.9 & 81.7 & 54.9 & 79.7 & \textbf{74.1} & 52.8 & 74.2 & 70.7 \\
    Liger-GLA-Mistral-7B & 0.02 & 80.1 & 78.7 & 49.3 & 76.3 & 70.1 & 36.3 & 70.9 & 65.1 \\
    Liger-GLA-Llama-3-8B & 0.02 & 80.3 & 81.1 & 52.5 & 76.3 & 72.0 & 43.4 & 72.4 & 67.6 \\
    \midrule
    \multicolumn{10}{l}{\textit{Linearized (Unbounded)}} \\
    Mamba2-LLaMA-3-8B & 20 & 76.8 & 74.1 & 48.0 & 70.8 & 58.6 & 43.2 & 65.6 & 61.9 \\
    \textbf{Mistral-7B-Lizard (Ours)} & 0.04 & 81.8 & 83.2 & 55.8 & \textbf{79.8} & 72.0 & 60.8 & 74.5 & 72.2 \\
    \textbf{LLaMA-3-8B-Lizard (Ours)} & 0.04 & \textbf{82} & \textbf{83.5} & \textbf{56.7} & 79.3 & 71.7 & \textbf{61.2} & \textbf{74.6} & \textbf{72.4} \\
    \bottomrule
  \end{tabular}
  \caption{
    Performance comparison of Lizard and existing 7B-size subquadratic LLMs. Linearized models are categorized as \textit{Bounded} (limited to context length) or \textit{Unbounded} (capable of extrapolating to longer sequences). 
    }
  \label{tab:main-benchmark}
\end{table*}

The core of the Lizard framework is the replacement of the softmax attention layer with an augmented, hybrid subquadratic mechanism.  This transformation is achieved through a two-stage training process: an initial attention approximation stage to mimic the teacher model, followed by a fine-tuning stage to align the new architecture with downstream language modeling objectives.

\subsection{First Stage: Approximating Softmax Attention for Unbounded Context}
In the teacher model, query and key vectors are transformed by a RoPE module before the attention computation. The full RoPE-infused softmax attention output, which we aim to approximate, is:
\begin{equation*}
\mathbf{y}_i = \sum_{t=1}^{i}
\frac{
\exp\left(\varphi_R(\mathbf{q}_i)^\top \varphi_R(\mathbf{k}t) / \sqrt{d}\right)
}{
\sum_{j=1}^{i}
\exp\left(\varphi_R(\mathbf{q}_i)^\top \varphi_R(\mathbf{k}_j) / \sqrt{d}\right)
} \mathbf{v}_t
\end{equation*}
where $\varphi_R(.)$ denotes the RoPE transformation. By training our RoPE-free mechanism to replicate this output, we distill both the attention patterns and the positional awareness of the teacher model.

\noindent \textbf{Learnable Gating for Adaptive Memory Control and Length Extrapolation:}
To solve the core limitations of prior work, we augment the linear attention mechanism with a learnable gating module, forming a data-adaptive recurrent structure. The output of the resulting RoPE-free Gated Linear Attention is computed as:
\begin{equation*}
\hat{\mathbf{y}}_i =
\frac{
\phi_q(\mathbf{q}_i)^\top \left( \sum_{t=1}^{i} \textcolor{red}{\left(\prod_{l=t+1}^{i} \mathbf{\Gamma}_l\right)} \phi_k(\mathbf{k}_t) \mathbf{v}_t^\top \right)
}{
\phi_q(\mathbf{q}_i)^\top \left( \sum_{j=1}^{i} \textcolor{red}{\left(\prod_{l=j+1}^{i} \mathbf{\Gamma}_l\right)} \phi_k(\mathbf{k}_j) \right)
}
\end{equation*}
where $\mathbf{\Gamma}_i = \text{sigmoid}(\textcolor{red}{\mathbf{W}_{\gamma}} \mathbf{x}_i)$ is the learnable gating factor. The gating mechanism plays a dual role in the attention transformation. First, it implicitly captures relative positional information by controlling the decay of past contributions. Unlike RoPE, which relies on predefined sinusoidal patterns, the data-adaptive gating factors enable better generalization across context lengths. Second, the gating mechanism provides adaptive memory control by allowing the model to dynamically determine how much past information to retain or forget. This property supports a recurrent formulation that enables constant-memory inference through an incremental hidden state update $\mathbf{S}_i$, which summarizes the historical information up to position $i$:
\[
\mathbf{S}_i = \mathbf{\Gamma}_i \mathbf{S}_{i-1} + \phi_k(\mathbf{k}_i) \mathbf{v}_i^\top, \quad
\hat{\mathbf{y}}_i = \phi_q(\mathbf{q}_i)^\top \mathbf{S}_i
\]
This state update removes the need to store the full key-value sequence, allowing constant-memory inference. 

\noindent \textbf{Anchor Window Attention for Local Precision:}
While the gated recurrent structure excels at compressing global context, it can lose the sharp, spiky detail of softmax attention. To preserve this local precision, we combine the globally-aware GLA with Anchor Window Attention. This mechanism augments a local sliding window with a set of $m$ learnable meta-memory tokens $\textcolor{red}{\mathbf{t}} \in \mathbb{R}^m$. Conceptually, these tokens function similarly to soft prompts, but mathematically they act as dynamic bias terms in the denominator of the attention computation.

These tokens are engineered to function as dedicated attention sinks, whose primary role is to stabilize the attention distribution by absorbing attention weight, without directly contributing their value vectors to the final output. This allows the model to divert attention mass to these sinks when local information is less relevant, thereby managing the massive activations phenomenon \citep{sun2024massive, gu2025when} and preserving the fidelity of the local context. To achieve this, we modify the standard softmax computation. The output at position $i$ is computed as:
\begin{equation*}
\hat{\mathbf{y}}_i = \frac{
\sum_{t=i-w+1}^{i} \exp(\mathbf{q}_i^\top \mathbf{k}_t / \sqrt{d}) \mathbf{v}_t
}{
\sum_{j=0}^{m-1} \textcolor{red}{\mathbf{t}_j} + \sum_{t=i-w+1}^{i} \exp(\mathbf{q}_i^\top \mathbf{k}_t / \sqrt{d})
}
\end{equation*}
where $\textcolor{red}{\mathbf{t}_j}$ is a learnable scalar parameter representing the logit of a meta-memory token. This formulation allows the model to manage powerful global signals via the meta-memory sinks while focusing the output computation on the fine-grained local context. This is achieved while maintaining a fixed-size key-value cache of $w+m$ tokens for constant-memory inference.

\begin{table*}[t]
  \centering
  \small
  \renewcommand{\arraystretch}{0.8}
  \setlength{\tabcolsep}{0.8pt}
  \begin{tabular}{lcccccccccc}
    \toprule
    \textbf{Model} & \textbf{Training} & \textbf{PiQA} & \textbf{ARC-e} & \textbf{ARC-c} & \textbf{Hella.} & \textbf{Wino.} & \textbf{MMLU} & \textbf{Avg.} & \textbf{Avg.} \\
    & \textbf{Tokens (B)} & acc & acc & acc\_norm & acc\_norm & acc & (5-shot) & (no MMLU) & \\
    \midrule
    LLaMA-3-8B & 15000 & 79.9 & 80.1 & 53.3 & 73.1 & 79.1 & 66.6 & 73.1 & 72.0 \\
    \midrule
    \multicolumn{10}{l}{\textit{Hybrid Softmax}} \\
    StripedHyena-Nous-7B & --    & 78.8 & 77.2 & 40.0 & 66.4 & 76.4 & 26.0 & 67.8 & 60.8 \\
    Zamba-7B & 1000  & 81.4 & 74.5 & 46.6 & 76.4 & 80.2 & 57.7 & 71.8 & 69.5 \\
    \midrule
    \multicolumn{10}{l}{\textit{Linearized (Keep 50\% Full Attn.) }} \\
    Mamba2-LLaMA-3 & 20 & 81.5 & 78.8 & 58.2 & 71.5 & 79.5 & 56.7 & 73.9 & 71.0 \\
    LLaMA-3-8B-Lizard (Ours) & 0.04 & \textbf{82.2} & \textbf{83.1} & \textbf{55.9} & \textbf{73.6} & \textbf{81.4} & \textbf{65.1} & \textbf{75.2} & \textbf{73.5} \\
    \bottomrule
  \end{tabular}
  \caption{
    Comparison of hybrid softmax models on language understanding benchmarks.
    }
  \label{tab:main-benchmark-hybrid}
\end{table*}

\noindent \textbf{Attention Approximation:}
We approximate the full softmax attention output, $\mathbf{Y}_{\text{softmax}}$, by combining the outputs of the globally-aware Gated Linear Attention and the locally-precise Anchor Window Attention. The final output, $\hat{\mathbf{Y}}_{lizard}$, is a combination of the two:
\begin{equation*}
\hat{\mathbf{Y}}_{lizard} = \hat{\mathbf{Y}}_{\text{gate}} + \alpha \cdot \hat{\mathbf{Y}}_{\text{anchor}}
\end{equation*}
where $\hat{\mathbf{Y}}_{\text{gate}}$ and $\hat{\mathbf{Y}}_{\text{anchor}}$ are the outputs from the gated and anchor window mechanisms. The learnable parameters are optimized by minimizing the discrepancy between our approximation and the teacher model's original attention output:
\begin{equation*}
\mathcal{L}_{\text{MSE}}(\phi, \mathbf{W}_\gamma, \mathbf{t}) =
\frac{1}{N} \sum_{l=1}^N
\left\| \mathbf{Y}_{\text{softmax}}^l - \hat{\mathbf{Y}}^l_{lizard} \right\|_F^2
\end{equation*}
where $N$ is the number of attention layers in the model. Overall, Lizard achieves an $O\bigl(L(w + m)d + Ld^2\bigr)$ time and space complexity. For generation, Lizard requires only $O((w + m)^2 d + d^2)$ time and space for every token.

\subsection{Second Stage: Aligning with Language Modeling}
While the first stage ensures a high-fidelity architectural approximation, the second stage aligns the model with the downstream language modeling task. In this stage, the original softmax attention layers are replaced by Lizard attention layers, and the entire model is fine-tuned using the standard autoregressive language modeling objective:
$\mathcal{L}_{\text{LM}}(\theta) = -\sum_{i=1}^{L} \log P(x_i \mid x_{<i})$
This step bridges the gap between structural mimicry and optimal task performance, adapting the linearized model to its end-to-end objective.

\section{Hardware-Aware Algorithm for Efficient Training}

To achieve maximum efficiency, Gated Linear Attention (GLA) must be computed in a parallel form on hardware accelerators like GPUs. However, the standard parallel formulation suffers from a critical numerical instability that prevents the use of low-precision formats, thereby creating a performance bottleneck. 
The parallel form of Gated Linear Attention is expressed as:
\begin{equation*}
\hat{\mathbf{Y}}_{\text{gate}} = \left( \left(\left( \phi(\mathbf{Q}) \odot \mathbf{C} \right)
\left( \frac{\phi(\mathbf{K})}{\mathbf{C}} \right)^\top\right)
\odot \mathbf{M} \right) \mathbf{V}
\end{equation*}
where $\mathbf{C}$ is the matrix of cumulative gate products, with each row $\mathbf{c}_t = \prod_{j=1}^{t} \mathbf{\Gamma}_j$.

The matrix form of gated linear attention is not numerically stable, as the cumulative product of gating values $\mathbf{c}_t$ can become extremely small, leading to underflow and instability during training for the low precision format such as \texttt{bfloat16}. This forces a fallback to full-precision (\texttt{float32}) operations, which is $2-3 \times$ slower and memory intensive, preventing the use of hardware accelerators like Tensor Cores that are optimized for low-precision arithmetic.

We leverage the strictly non-negative property of the \textit{Hedgehog} feature map \citep{zhang2402hedgehog}, $\phi(\mathbf{x}) = [ \exp(\mathbf{x}\mathbf{W}) \oplus \exp(-\mathbf{x}\mathbf{W}) ]$. This exponential-based structure is critical, as it permits a stable reparameterization of the attention computation in log-space. We absorb the cumulative gate term $\mathbf{C}$ directly into the query and key projections, resulting in the following hardware-efficient formulation:
\begin{align*}
    \widetilde{\mathbf{Q}} &= \left[ \exp(\mathbf{Q}\mathbf{W} + \log \mathbf{C}) \oplus \exp(-\mathbf{Q}\mathbf{W} + \log \mathbf{C}) \right] \\
    \widetilde{\mathbf{K}} &= \left[ \exp(\mathbf{K}\mathbf{W} - \log \mathbf{C}) \oplus \exp(-\mathbf{K}\mathbf{W} - \log \mathbf{C}) \right]
\end{align*}
By shifting the unstable gating contributions into the feature space, this approach transforms the core operation into a standard General Matrix Multiplication (\textbf{GEMM}): $\hat{\mathbf{Y}}_{\text{gate}} = \left( \left( \widetilde{\mathbf{Q}} \widetilde{\mathbf{K}}^\top \right) \odot \mathbf{M} \right) \mathbf{V}$ which aligns with the native \texttt{mma.sync} Tensor Core instruction. This avoids custom CUDA kernels and leverages highly optimized GEMM routines in cuBLAS and FlashLinearAttention backends \citep{yang2024fla}, avoids a full-precision fallback.

To empirically validate the benefits of our hardware-aware reparameterization, we benchmark the forward-pass latency of the Lizard kernel against the standard Gated Linear Attention (GLA) implementation \citep{yang2024fla} across various batch sizes ($B$) and sequence lengths ($L$). As shown in Table~\ref{tab:hardware_speedup}, by shifting the gating contributions into the feature space and enabling native Tensor Core utilization, Lizard achieves a consistent 32\% to 36\% speedup while avoiding the precision fallback typically required by standard GLA kernels.

\begin{table}[h]
\centering
\small
\begin{tabular}{lccc}
\toprule
\textbf{Configuration} & \textbf{GLA} & \textbf{Lizard} & \textbf{Speedup} \\
\midrule
$B = 16, L = 2048$ & 4.29 ms & 3.25 ms & +32\% \\
$B = 16, L = 4096$ & 8.77 ms & 6.50 ms & +35\% \\
$B = 16, L = 8192$ & 18.90 ms & 13.80 ms & +36\% \\
$B = 32, L = 8192$ & 36.74 ms & 27.22 ms & +36\% \\
\bottomrule
\end{tabular}
\caption{Forward pass latency comparison on an NVIDIA A100-80GB GPU. Lizard delivers significant speedups by enabling efficient low-precision arithmetic on hardware accelerators.}
\label{tab:hardware_speedup}
\end{table}

\section{Experimental Study}
\label{sec:experiments}

\begin{figure*}
  \centering
  \includegraphics[width=\linewidth]{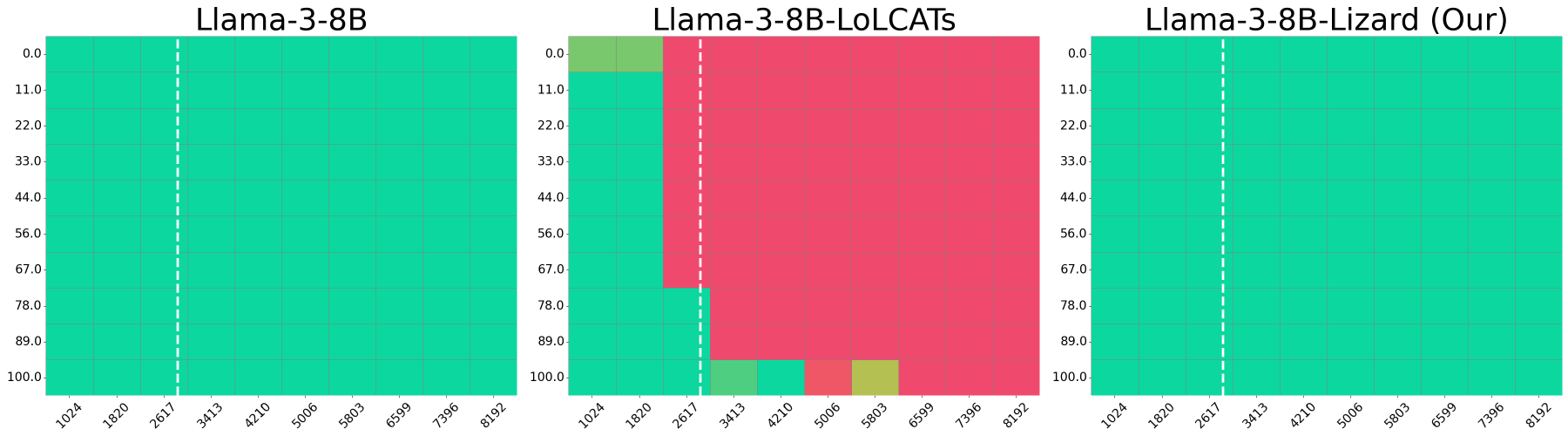}
  \caption{\textbf{Needle-in-a-Haystack evaluation.}  Each cell shows retrieval accuracy by sequence length (X-axis) and target distance (Y-axis). Green indicates success; red indicates failure. The white dashed line marks the max training length.}
  \label{fig:needle}
\end{figure*}

In this section, we present our experimental results, focusing on three key aspects:
\begin{enumerate}
    \item \textbf{Language Modeling Benchmarks:} We evaluate Lizard on standard language modeling datasets and compare its performance against existing subquadratic alternatives and linearizations. Our results indicate that Lizard matches the average performance of the teacher model and significantly outperforms other baselines by a large margin. 

    \item \textbf{Long-Context Associative Recall:} We evaluate the model's retrieval capabilities across extreme sequence lengths using the Needle-in-a-Haystack and RULER benchmarks. Unlike prior linearization methods that fail immediately beyond their training context, Lizard demonstrates robust length extrapolation up to 64K tokens. We show that our hybrid configuration achieves perfect retrieval accuracy and near-lossless recovery of the teacher model's performance on the 5-shot MMLU benchmark, effectively bridging the gap between efficiency and precision.
    
    \item \textbf{Generation Efficiency:} We compare the generation throughput of Lizard with the teacher model equipped with FlashAttention-2. While the teacher model quickly runs out of memory at a sequence length of 32K, Lizard maintains constant memory usage and throughput, enabling efficient generation with infinite context.

    \item \textbf{Architectural and Ablation Analysis:} We conduct a detailed analysis of Lizard's design choices, including the structure of the gating module and the contribution of each component. These studies help identify the most effective configurations for performance and efficiency.
\end{enumerate}

\noindent \textbf{Experimental Setup:} We conduct our experiments using two widely used Transformer-based LLMs: Mistral-7B \citep{jiang263830494mistral} and Llama-3-8B \citep{grattafiori2024llama} as teacher models. For training, we utilize a curated subset of 50K high-quality examples cleaned Alpaca dataset \footnote{https://huggingface.co/datasets/yahma/alpaca-cleaned} \citep{alpaca}. By default, we use a scalar gating structure, where $\Gamma_i = \gamma_i \mathbf{1}_d^\top, \gamma_i = \sigma(W_{\gamma} \mathbf{x}_i)$, with $\gamma_i \in \mathbb{R}$ and $W_{\gamma} \in \mathbb{R}^{d \times 1}$. We also explore various gating module designs, which are discussed in Section~\ref{sec:ablation}.

In the \emph{First Stage}, we train the feature maps $\phi_q$, $\phi_k$, and the gating parameter matrix $W_{\gamma}$ jointly to approximate full softmax attention and RoPE patterns. For the sliding window attention module, we use a small window size of $w = 128$ and $m = 4$ meta tokens. In the \emph{Second Stage}, we employ Low-Rank Adaptation (LoRA)~\citep{hu2022lora} for parameter-efficient fine-tuning. Specifically, LoRA is applied to the projection matrices \(W_Q\), \(W_K\), and \(W_V\), with a default rank \(r = 8\) and scaling factor \(\alpha = 16\). Both stages are trained for 2 epochs, corresponding to 20M tokens per stage. We use the AdamW optimizer and a cosine learning rate schedule with a 10\% linear warmup. Training is performed using Fully Sharded Data Parallelism (FSDP-2) \citep{zhao2023pytorch} across 8$\times$A100 80GB GPUs. The peak learning rate is set to $1 \times 10^{-3}$ for the first stage and $5 \times 10^{-4}$ for the second stage. We adopt a global batch size of 8, with each example having a maximum sequence length of 2048 tokens.

\subsection{Language Modeling Benchmarks}

\begin{table*}[t]
\centering
\small
\resizebox{\textwidth}{!}{%
\begin{tabular}{cccc}
\toprule
\textbf{Model} & \textbf{Gating Parameterization} & \textbf{Learnable Parameters} & \textbf{MMLU 5-shot} \\
\midrule
\textbf{Lizard (Ours)} & $\Gamma_i = \gamma_i \mathbf{1}_d^\top,\; \gamma_i = \sigma(W_{\gamma} \mathbf{x}_i)$ & 
$W_{\gamma} \in \mathbb{R}^{d \times 1}$ & 61.2 \\

Mamba-2~\citep{dao2024transformers} & $\Gamma_i = \gamma_i \mathbf{1}_d^\top,\; \gamma_i = \exp\left( -\operatorname{softplus}(\mathbf{x}_i W_\gamma) \cdot \exp(a) \right)$ & 
$W_{\gamma} \in \mathbb{R}^{d \times 1},\; a \in \mathbb{R}$ & 57.6 \\

GLA~\citep{yang2024gated} & $\Gamma_i = \sigma\left(\mathbf{x}_i W_{\gamma_1} W_{\gamma_2} \right)$ & 
$W_{\gamma_1} \in \mathbb{R}^{d \times 16},\; W_{\gamma_2} \in \mathbb{R}^{16 \times d}$ & 53.5 \\
1D-Pooling & $\Gamma_i = \sigma(Pooling(\textbf{k}_t))$ & 
N/A  & 44.1 \\
\bottomrule
\end{tabular}%

}
\caption{Performance comparison of different gating designs and their parameterizations.}
\label{tab:formulations}
\end{table*}

We evaluate Lizard on six popular language understanding benchmarks from the LM Evaluation Harness (LM Eval) \footnote{https://github.com/EleutherAI/lm-evaluation-harness} \citep{eval-harness}, including PiQA~\citep{bisk2020piqa}, ARC-easy (ARC-e) and ARC-challenge (ARC-c)~\citep{clark2018think}, HellaSwag (Hella.)~\citep{zellers-etal-2019-hellaswag}, WinoGrande (Wino.)~\citep{sakaguchi2021winogrande}, and MMLU~\citep{hendrycks2020measuring}. Notably, Lizard is able to closely recover the performance of the teacher model and achieves near-lossless accuracy across tasks in average, demonstrating that it preserves the original model’s language understanding capabilities.

We compare Lizard against two groups of baselines. The first group, presented in Table~\ref{tab:main-benchmark}, consists of \textbf{subquadratic LLMs}, including models pre-trained from scratch with linear or subquadratic attention mechanisms, such as Mamba~\citep{gu2023mamba}, RWKV-6~\citep{peng-etal-2023-rwkv}, TransNormerLLM-7B~\citep{qin2307transnormerllm}, Hawk and Griffin~\citep{de2402griffin}, as well as linearized variants such as SUPRA~\citep{mercat2024linearizing}, Mamba2LLaMA~\citep{wang2024mamba}, LoLCATs~\citep{zhang2025lolcats}, Liger~\citep{lan2025liger}. Lizard consistently outperforms prior approaches, particularly on the 5-shot MMLU benchmark, where it achieves an 18\% improvement over previous methods with similar extrapolation capabilities. Compared to LoLCATs and Liger, which does not generalize beyond training context, Lizard scores 9.4 and 24.5 points higher on the 5-shot MMLU, respectively.

The second group shown in Table~\ref{tab:main-benchmark-hybrid}, \textbf{Hybrid Softmax Architectures} includes models that combine full softmax with subquadratic attention layers. We compare with models such as StripedHyena-Nous-7B \citep{poli2023stripedhyena} and Zamba-7B \citep{glorioso2024zamba}. Following the same configuration of Mamba2-LLaMA-3-8B~\citep{wang2024mamba}, which retains 50\% softmax layers. On 5-shot MMLU, Lizard score $65.1$, closely matching the 66.1 score of the original LLaMA-3-8B teacher model, while outperforming all hybrid baselines.

\begin{figure}
    \centering
    \includegraphics[width=\linewidth]{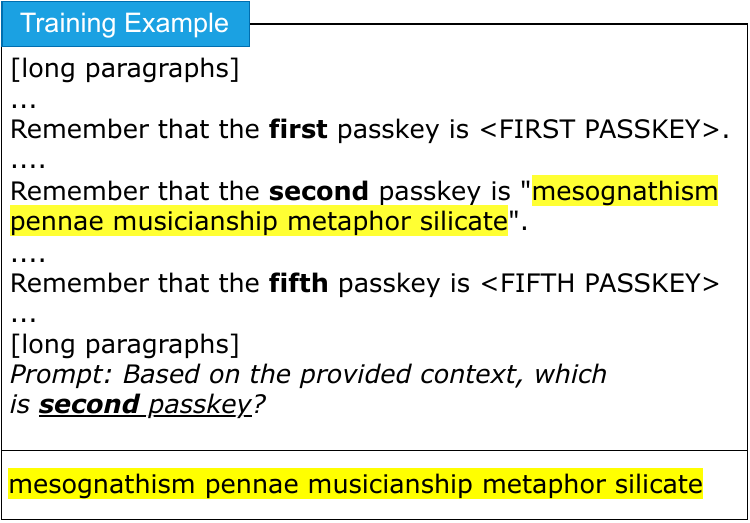}
    \caption{Example from the synthetic passkey retrieval dataset.}
    \label{fig:passkey}
\end{figure}
\subsection{Recall Evaluations}

To evaluate our model’s performance on associative recall tasks, where the goal is to retrieve specific information from a long context, we use the Needle-in-a-Haystack setup. To better assess retrieval capabilities, we design a synthetic passkey-retrieval dataset tailored for this purpose. As illustrated in Figure~\ref{fig:passkey}, each input sequence contains five randomly generated passkeys, each of length 5-8 tokens, inserted at random positions within a long sequence. At test time, the model is prompted to retrieve one selected passkey from the five embedded within the sequence. We generate 10,000 synthetic examples, train the model on sequences of length 2048, and evaluate its performance on longer sequences ranging from 2048 to 8192 tokens to assess its generalization and recall capabilities in long-context settings.

\noindent \textbf{NIAH Benchmark:} We evaluate Lizard and compare its performance against the teacher model and LoLCATs \citep{zhang2025lolcats}, a recent state-of-the-art linearization method. Figure ~\ref{fig:needle} reports the results of all three models on the associative recall test set. We find that Lizard significantly outperforms LoLCATs in both associative recall accuracy and length generalization.
Notably, Lizard is able to perfectly retrieve the correct passkey across various context lengths, while LoLCATs fails when the sequence length exceeds the training window. This highlights the strength of the gated recurrent structures, which effectively compresses global context and does not rely solely on the expressiveness of local sliding window attention. 

\noindent \textbf{RULER Benchmark:} To evaluate long-range recall beyond synthetic passkey retrieval, we benchmark our strongest configuration (Hybrid Lizard, 50\% softmax replacement) on the RULER benchmark \citep{hsieh2024ruler} using LLaMA-3-8B-Instruct as the teacher. As shown in Table~\ref{tab:ruler}, Lizard maintains high accuracy up to 32K context lengths, closely matching the retrieval performance of the full-attention teacher model. This confirms that Lizard effectively preserves the model's ability to handle complex, long-context dependencies without the quadratic memory overhead of standard Transformers.

\begin{table}[h]
\centering
\small
\begin{tabular}{lcccc}
\toprule
\textbf{Model} & \textbf{4K} & \textbf{8K} & \textbf{16K} & \textbf{32K} \\
\midrule
LLaMA-3-8B (Teacher) & 92.3 & 90.5 & 85.7 & 80.5 \\
Lizard (Hybrid) & 92.5 & 91.2 & 85.2 & 81.3 \\
\bottomrule
\end{tabular}
\caption{Retrieval accuracy on the RULER benchmark. Hybrid Lizard maintains performance parity with the teacher model across scaling context lengths.}
\label{tab:ruler}
\end{table}

\subsection{Generation Efficiency}

\begin{figure*}
  \centering
  \includegraphics[width=0.95\linewidth]{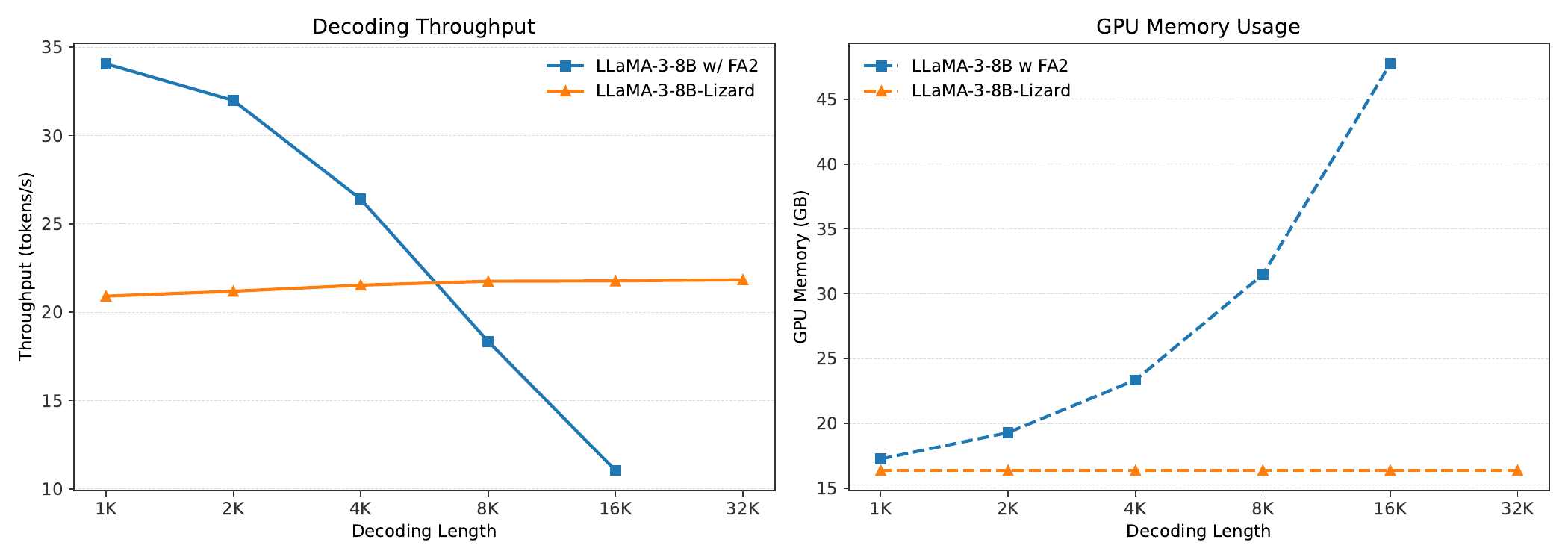}
  \caption{Throughput and memory comparison.}
  \label{fig:efficiency}
\end{figure*}
We assess the efficiency of Lizard by comparing its throughput and memory usage to that of the teacher model across input sequence lengths from 1K to 32K, using a batch size of 16. As shown in Figure~\ref{fig:efficiency}, the teacher model with FlashAttention-2 \citep{dao2023flashattention} encounters out-of-memory (OOM) issues at longer sequence lengths. In contrast, Lizard maintains constant memory consumption and stable throughput throughout.
All experiments were conducted on an NVIDIA A100 80GB GPU.

\subsection{Ablation and Architectural Analysis} 
\label{sec:ablation}

\noindent \textbf{Impact of Architectural Components:} We conduct an ablation study to evaluate the contribution of each individual module in Lizard. As shown in Table~\ref{tab:ablation_mmlu_avg}, removing the Sliding Window Attention (SWA) or the gated recurrent module results in a catastrophic performance drop on the 5-shot MMLU benchmark, with scores falling to 39.7 and 42.2, respectively. Furthermore, omitting the initial attention approximation stage significantly hinders the model's ability to recover the teacher's reasoning capabilities (50.8 MMLU). Notably, we observe that our default LoRA-based fine-tuning performs nearly as well as full fine-tuning (61.2 vs. 61.4), validating our choice of parameter-efficient adaptation for the linearization process.

\begin{table}[h]
\centering
\small
\begin{tabular}{lc}
\toprule
\textbf{Model Configuration} & \textbf{MMLU (5-shot)}\\
\midrule
LLaMA-3-8B-Lizard (Full) & \textbf{61.2} \\
\midrule
-- w/o Sliding Window Attention (SWA) & 39.7 \\
-- w/o Gated Module & 42.2 \\
-- w/o Attention Approximation & 50.8  \\ 
-- Full Fine-tuning (No LoRA) & 61.4 \\
\bottomrule
\end{tabular}
\caption{Ablation results on LLaMA-3-8B demonstrating the necessity of each architectural component for performance recovery.}
\label{tab:ablation_mmlu_avg}
\end{table}

\noindent \textbf{Gated Structures Design} Table~\ref{tab:formulations} presents a comparison of different gating designs and parameterizations based on recent architectural advances. We experiment with multiple formulations, ranging from minimal scalar gates to more expressive multi-layer projections. Our results show that the Lizard parameterization achieves the highest performance on the 5-shot MMLU benchmark. While complex gated recurrent structures offer greater modeling flexibility, we observe that their effectiveness is limited by the need to initialize these modules from scratch. Heavier parameterizations can lead to overfitting or instability during fine-tuning, ultimately degrading performance. In contrast, lightweight designs with minimal additional parameters are easier to train and generalize better, resulting in stronger overall performance. Additionally, we evaluate a pooling-based variant where the gating values are derived from 1D pooling over key vectors, eliminating the need for any learnable parameters. However, this configuration results in a significant drop in performance. This suggests that having a learnable gating mechanism, even with minimal parameters, is crucial for capturing meaningful temporal patterns and maintaining strong performance on downstream tasks.

\begin{table}[h]
\centering
\small
\begin{tabular}{lcccc}
\toprule
\textbf{} & $w=32$ & $w=64$ & $w=128$ & \textbf{$w=256$} \\
\midrule
$m=2$ & 51.3 & 54.1 & 58.6 & 42.2 \\
$m=4$ & 52.4 & 57.6 & \textbf{61.2} & 44.6 \\
$m=6$ & 52.4 & 57.9 & 60.8 & 43.8 \\
\bottomrule
\end{tabular}
\caption{5-shot MMLU performance with varying window and meta token sizes}
\label{tab:mmlu_meta_window}
\end{table}

\noindent \textbf{Effect of window and meta memory size} Table~\ref{tab:mmlu_meta_window} presents an ablation study evaluating the impact of varying the local attention window size ($w$) and the number of meta tokens ($m$) on 5-shot MMLU performance. While increasing the window size can improve performance, it does not guarantee consistent gains. For instance, performance peaks at $w=128$ for $m=4$, but drops significantly at $w=256$. 
We formalize this phenomenon as \textit{Local Attention Dominance}. During the joint training of the hybrid attention mechanism, the sharp, spiky distributions produced by the exact local softmax can yield massive gradient signals. If the local window is excessively large, these local gradients dominate the optimization process. Consequently, the model becomes overly reliant on the local sliding window, and the recurrent module fails to effectively learn the gating dynamics required for global context compression. This highlights a optimization trade-off: tightly constraining the local window is not merely a computational optimization, but a necessary structural constraint to force the model to utilize its global recurrent memory.

\begin{table}[h]
\centering
\small
\begin{tabular}{lcc}
\toprule
\textbf{LoRA Rank} & \textbf{MMLU (5-shot)} & \textbf{Avg. (no MMLU)} \\
\midrule
4     & 59.7 & 74.1 \\
\textbf{8}     & \textbf{61.2} & \textbf{74.6} \\
16    & 60.6 & 73.3 \\
32    & 61.0 & 74.5 \\
64    & 59.2 & 74.0 \\
\bottomrule
\end{tabular}
\caption{Effect of LoRA Rank on LLaMA-3-8B-Lizard.}
\label{tab:lora_rank_ablation}
\end{table}

\noindent \textbf{Effect of LoRA Rank:} We investigate the impact of the LoRA rank ($r$) on model performance to determine the minimum parameter overhead required for successful linearization. As shown in Table~\ref{tab:lora_rank_ablation}, a rank of 8 is sufficient to achieve peak performance, matching or even slightly surpassing full fine-tuning on both 5-shot MMLU and the average across tasks. Interestingly, increasing the rank beyond 8 does not yield further gains and, in some cases, leads to slight performance degradation, likely due to overfitting on the relatively small linearization dataset.

\section{Conclusion}
In this work, we introduced {\bf Lizard}, a novel linearization framework designed to bridge the gap between the high performance of Transformer-based LLMs and the computational efficiency of subquadratic architectures. Our extensive evaluations demonstrate that Lizard achieves near-lossless recovery of teacher performance, significantly outperforming prior linearization methods. Notably, Lizard exhibits superior associative recall and maintains high retrieval accuracy. Lizard provides a scalable path for transforming existing state-of-the-art LLMs into efficient, constant-memory inference engines without sacrificing the reasoning capabilities developed during massive-scale pretraining. We believe this framework offers a practical solution for deploying advanced language models in resource-constrained.

\section{Limitations}
\label{sec:limitations}

Despite the promising performance and efficiency gains demonstrated by Lizard, our approach has two key limitations. First, Lizard still relies on a strong pretrained backbone to achieve high quality. As with many recent distillation-based or hybrid architectures, the success of our method depends heavily on the expressiveness and generalization capacity of the teacher model. Without access to a high-quality pretrained model (e.g., Llama-3-8B), the performance of Lizard may degrade significantly, especially on complex reasoning and multilingual tasks. Second, Lizard inherits the inherent tradeoffs present in linear attention mechanisms. While our design enables constant-time and constant-memory inference with infinite context length, it still exhibits a recall-memory tradeoff. That is, models with fixed-size state representations, such as our gated linear attention, may underperform in recall-intensive tasks compared to full attention models, which maintain a growing key-value cache. This aligns with recent findings that efficient alternatives to attention often struggle to retain long-range information critical for grounding generations in earlier context. As a result, while Lizard expands the throughput-recall Pareto frontier, it does not eliminate the tradeoff entirely.

\section*{Acknowledgments}

This research was partially supported by NSF Grant \#2239570. This research is also supported in part by the Office of the Director of National Intelligence (ODNI), Intelligence Advanced Research Projects Activity (IARPA), via the HIATUS Program contract 2022-22072200003. The views and conclusions contained herein are those of the authors and should not be interpreted as necessarily representing the official policies, either expressed or implied, of ODNI, IARPA, or the U.S. Government.

\bibliography{custom}

\begin{thebibliography}{39}
\providecommand{\natexlab}[1]{#1}

\bibitem[{Achiam et~al.(2023)Achiam, Adler, Agarwal, Ahmad, Akkaya, Aleman,
  Almeida, Altenschmidt, Altman, Anadkat et~al.}]{achiam2023gpt}
Josh Achiam, Steven Adler, Sandhini Agarwal, Lama Ahmad, Ilge Akkaya,
  Florencia~Leoni Aleman, Diogo Almeida, Janko Altenschmidt, Sam Altman,
  Shyamal Anadkat, and 1 others. 2023.
\newblock Gpt-4 technical report.
\newblock \emph{arXiv preprint arXiv:2303.08774}.

\bibitem[{Adnan et~al.(2024)Adnan, Arunkumar, Jain, Nair, Soloveychik, and
  Kamath}]{adnan2024keyformer}
Muhammad Adnan, Akhil Arunkumar, Gaurav Jain, Prashant~J Nair, Ilya
  Soloveychik, and Purushotham Kamath. 2024.
\newblock Keyformer: Kv cache reduction through key tokens selection for
  efficient generative inference.
\newblock \emph{Proceedings of Machine Learning and Systems}, 6:114--127.

\bibitem[{Bisk et~al.(2020)Bisk, Zellers, Gao, Choi et~al.}]{bisk2020piqa}
Yonatan Bisk, Rowan Zellers, Jianfeng Gao, Yejin Choi, and 1 others. 2020.
\newblock Piqa: Reasoning about physical commonsense in natural language.
\newblock In \emph{Proceedings of the AAAI conference on artificial
  intelligence}, volume~34, pages 7432--7439.

\bibitem[{Clark et~al.(2018)Clark, Cowhey, Etzioni, Khot, Sabharwal, Schoenick,
  and Tafjord}]{clark2018think}
Peter Clark, Isaac Cowhey, Oren Etzioni, Tushar Khot, Ashish Sabharwal, Carissa
  Schoenick, and Oyvind Tafjord. 2018.
\newblock Think you have solved question answering? try arc, the ai2 reasoning
  challenge.
\newblock \emph{arXiv preprint arXiv:1803.05457}.

\bibitem[{Dao(2023)}]{dao2023flashattention}
Tri Dao. 2023.
\newblock Flashattention-2: Faster attention with better parallelism and work
  partitioning.
\newblock \emph{arXiv preprint arXiv:2307.08691}.

\bibitem[{Dao and Gu(2024)}]{dao2024transformers}
Tri Dao and Albert Gu. 2024.
\newblock Transformers are ssms: Generalized models and efficient algorithms
  through structured state space duality.
\newblock \emph{arXiv preprint arXiv:2405.21060}.

\bibitem[{De et~al.()De, Smith, Fernando, Botev, Cristian-Muraru, Gu, Haroun,
  Berrada, Chen, Srinivasan et~al.}]{de2402griffin}
Soham De, Samuel~L Smith, Anushan Fernando, Aleksandar Botev, George
  Cristian-Muraru, Albert Gu, Ruba Haroun, Leonard Berrada, Yutian Chen,
  Srivatsan Srinivasan, and 1 others.
\newblock Griffin: Mixing gated linear recurrences with local attention for
  efficient language models, 2024.
\newblock \emph{URL https://arxiv. org/abs/2402.19427}, page~50.

\bibitem[{Gao et~al.(2024)Gao, Tow, Abbasi, Biderman, Black, DiPofi, Foster,
  Golding, Hsu, Le~Noac'h, Li, McDonell, Muennighoff, Ociepa, Phang, Reynolds,
  Schoelkopf, Skowron, Sutawika, Tang, Thite, Wang, Wang, and
  Zou}]{eval-harness}
Leo Gao, Jonathan Tow, Baber Abbasi, Stella Biderman, Sid Black, Anthony
  DiPofi, Charles Foster, Laurence Golding, Jeffrey Hsu, Alain Le~Noac'h,
  Haonan Li, Kyle McDonell, Niklas Muennighoff, Chris Ociepa, Jason Phang,
  Laria Reynolds, Hailey Schoelkopf, Aviya Skowron, Lintang Sutawika, and 5
  others. 2024.
\newblock \href {https://doi.org/10.5281/zenodo.12608602} {The language model
  evaluation harness}.

\bibitem[{Glorioso et~al.(2024)Glorioso, Anthony, Tokpanov, Whittington,
  Pilault, Ibrahim, and Millidge}]{glorioso2024zamba}
Paolo Glorioso, Quentin Anthony, Yury Tokpanov, James Whittington, Jonathan
  Pilault, Adam Ibrahim, and Beren Millidge. 2024.
\newblock Zamba: A compact 7b ssm hybrid model.
\newblock \emph{arXiv preprint arXiv:2405.16712}.

\bibitem[{Goldstein et~al.(2025)Goldstein, Alcaide, Lu, and
  Cheah}]{goldstein2025radlads}
Daniel Goldstein, Eric Alcaide, Janna Lu, and Eugene Cheah. 2025.
\newblock Radlads: Rapid attention distillation to linear attention decoders at
  scale.
\newblock \emph{arXiv preprint arXiv:2505.03005}.

\bibitem[{Grattafiori et~al.(2024)Grattafiori, Dubey, Jauhri, Pandey, Kadian,
  Al-Dahle, Letman, Mathur, Schelten, Vaughan et~al.}]{grattafiori2024llama}
Aaron Grattafiori, Abhimanyu Dubey, Abhinav Jauhri, Abhinav Pandey, Abhishek
  Kadian, Ahmad Al-Dahle, Aiesha Letman, Akhil Mathur, Alan Schelten, Alex
  Vaughan, and 1 others. 2024.
\newblock The llama 3 herd of models.
\newblock \emph{arXiv preprint arXiv:2407.21783}.

\bibitem[{Gu and Dao(2023)}]{gu2023mamba}
Albert Gu and Tri Dao. 2023.
\newblock Mamba: Linear-time sequence modeling with selective state spaces.
\newblock \emph{arXiv preprint arXiv:2312.00752}.

\bibitem[{Gu et~al.(2025)Gu, Pang, Du, Liu, Zhang, Du, Wang, and
  Lin}]{gu2025when}
Xiangming Gu, Tianyu Pang, Chao Du, Qian Liu, Fengzhuo Zhang, Cunxiao Du,
  Ye~Wang, and Min Lin. 2025.
\newblock \href {https://openreview.net/forum?id=78Nn4QJTEN} {When attention
  sink emerges in language models: An empirical view}.
\newblock In \emph{The Thirteenth International Conference on Learning
  Representations}.

\bibitem[{Hendrycks et~al.(2020)Hendrycks, Burns, Basart, Zou, Mazeika, Song,
  and Steinhardt}]{hendrycks2020measuring}
Dan Hendrycks, Collin Burns, Steven Basart, Andy Zou, Mantas Mazeika, Dawn
  Song, and Jacob Steinhardt. 2020.
\newblock Measuring massive multitask language understanding.
\newblock \emph{arXiv preprint arXiv:2009.03300}.

\bibitem[{Hsieh et~al.(2024)Hsieh, Sun, Kriman, Acharya, Rekesh, Jia, Zhang,
  and Ginsburg}]{hsieh2024ruler}
Cheng-Ping Hsieh, Simeng Sun, Samuel Kriman, Shantanu Acharya, Dima Rekesh, Fei
  Jia, Yang Zhang, and Boris Ginsburg. 2024.
\newblock Ruler: What's the real context size of your long-context language
  models?
\newblock \emph{arXiv preprint arXiv:2404.06654}.

\bibitem[{Hu et~al.(2022)Hu, Shen, Wallis, Allen-Zhu, Li, Wang, Wang, Chen
  et~al.}]{hu2022lora}
Edward~J Hu, Yelong Shen, Phillip Wallis, Zeyuan Allen-Zhu, Yuanzhi Li, Shean
  Wang, Lu~Wang, Weizhu Chen, and 1 others. 2022.
\newblock Lora: Low-rank adaptation of large language models.
\newblock \emph{ICLR}, 1(2):3.

\bibitem[{Jiang et~al.()Jiang, Sablayrolles, Mensch, Bamford, Chaplot,
  de~Las~Casas, Bressand, Lengyel, Lample, Saulnier
  et~al.}]{jiang263830494mistral}
AQ~Jiang, A~Sablayrolles, A~Mensch, C~Bamford, DS~Chaplot, D~de~Las~Casas,
  F~Bressand, G~Lengyel, G~Lample, L~Saulnier, and 1 others.
\newblock Mistral 7b, arxiv abs/2310.06825 (2023).
\newblock \emph{URL: https://api. semanticscholar. org/CorpusID}, 263830494.

\bibitem[{Katharopoulos et~al.(2020)Katharopoulos, Vyas, Pappas, and
  Fleuret}]{katharopoulos2020transformers}
Angelos Katharopoulos, Apoorv Vyas, Nikolaos Pappas, and Fran{\c{c}}ois
  Fleuret. 2020.
\newblock Transformers are rnns: Fast autoregressive transformers with linear
  attention.
\newblock In \emph{International conference on machine learning}, pages
  5156--5165. PMLR.

\bibitem[{Lan et~al.(2025)Lan, Sun, Hu, Du, and Cheng}]{lan2025liger}
Disen Lan, Weigao Sun, Jiaxi Hu, Jusen Du, and Yu~Cheng. 2025.
\newblock \href {https://openreview.net/forum?id=1PfZs0xC2v} {Liger:
  Linearizing large language models to gated recurrent structures}.
\newblock In \emph{Forty-second International Conference on Machine Learning}.

\bibitem[{Mann et~al.(2020)Mann, Ryder, Subbiah, Kaplan, Dhariwal, Neelakantan,
  Shyam, Sastry, Askell, Agarwal et~al.}]{mann2020language}
Ben Mann, N~Ryder, M~Subbiah, J~Kaplan, P~Dhariwal, A~Neelakantan, P~Shyam,
  G~Sastry, A~Askell, S~Agarwal, and 1 others. 2020.
\newblock Language models are few-shot learners.
\newblock \emph{arXiv preprint arXiv:2005.14165}, 1:3.

\bibitem[{Mercat et~al.(2024)Mercat, Vasiljevic, Keh, Arora, Dave, Gaidon, and
  Kollar}]{mercat2024linearizing}
Jean Mercat, Igor Vasiljevic, Sedrick~Scott Keh, Kushal Arora, Achal Dave,
  Adrien Gaidon, and Thomas Kollar. 2024.
\newblock \href {https://openreview.net/forum?id=soGxskHGox} {Linearizing large
  language models}.
\newblock In \emph{First Conference on Language Modeling}.

\bibitem[{Peng et~al.(2023{\natexlab{a}})Peng, Alcaide, Anthony, Albalak,
  Arcadinho, Biderman, Cao, Cheng, Chung, Derczynski, Du, Grella, Gv, He, Hou,
  Kazienko, Kocon, Kong, Koptyra, Lau, Lin, Mantri, Mom, Saito, Song, Tang,
  Wind, Wo{\'z}niak, Zhang, Zhou, Zhu, and Zhu}]{peng-etal-2023-rwkv}
Bo~Peng, Eric Alcaide, Quentin Anthony, Alon Albalak, Samuel Arcadinho, Stella
  Biderman, Huanqi Cao, Xin Cheng, Michael Chung, Leon Derczynski, Xingjian Du,
  Matteo Grella, Kranthi Gv, Xuzheng He, Haowen Hou, Przemyslaw Kazienko, Jan
  Kocon, Jiaming Kong, Bart{\l}omiej Koptyra, and 13 others.
  2023{\natexlab{a}}.
\newblock \href {https://doi.org/10.18653/v1/2023.findings-emnlp.936} {{RWKV}:
  Reinventing {RNN}s for the transformer era}.
\newblock In \emph{Findings of the Association for Computational Linguistics:
  EMNLP 2023}, pages 14048--14077, Singapore. Association for Computational
  Linguistics.

\bibitem[{Peng et~al.(2023{\natexlab{b}})Peng, Alcaide, Anthony, Albalak,
  Arcadinho, Biderman, Cao, Cheng, Chung, Grella et~al.}]{peng2023rwkv}
Bo~Peng, Eric Alcaide, Quentin Anthony, Alon Albalak, Samuel Arcadinho, Stella
  Biderman, Huanqi Cao, Xin Cheng, Michael Chung, Matteo Grella, and 1 others.
  2023{\natexlab{b}}.
\newblock Rwkv: Reinventing rnns for the transformer era.
\newblock \emph{arXiv preprint arXiv:2305.13048}.

\bibitem[{Poli et~al.(2023)Poli, Wang, Massaroli, Quesnelle, Carlow, Nguyen,
  and Thomas}]{poli2023stripedhyena}
Michael Poli, Jue Wang, Stefano Massaroli, Jeffrey Quesnelle, Ryan Carlow, Eric
  Nguyen, and Armin Thomas. 2023.
\newblock Stripedhyena: Moving beyond transformers with hybrid signal
  processing models.
\newblock \emph{GitHub repository}, 12.

\bibitem[{Qin et~al.()Qin, Li, Sun, Sun, Shen, Han, Wei, Lv, Luo, Qiao
  et~al.}]{qin2307transnormerllm}
Zhen Qin, Dong Li, Weigao Sun, Weixuan Sun, Xuyang Shen, Xiaodong Han, Yunshen
  Wei, Baohong Lv, Xiao Luo, Yu~Qiao, and 1 others.
\newblock Transnormerllm: A faster and better large language model with
  improved transnormer, 2024.
\newblock \emph{URL https://arxiv. org/abs/2307.14995}.

\bibitem[{Sakaguchi et~al.(2021)Sakaguchi, Bras, Bhagavatula, and
  Choi}]{sakaguchi2021winogrande}
Keisuke Sakaguchi, Ronan~Le Bras, Chandra Bhagavatula, and Yejin Choi. 2021.
\newblock Winogrande: An adversarial winograd schema challenge at scale.
\newblock \emph{Communications of the ACM}, 64(9):99--106.

\bibitem[{Su et~al.(2024)Su, Ahmed, Lu, Pan, Bo, and Liu}]{su2024roformer}
Jianlin Su, Murtadha Ahmed, Yu~Lu, Shengfeng Pan, Wen Bo, and Yunfeng Liu.
  2024.
\newblock Roformer: Enhanced transformer with rotary position embedding.
\newblock \emph{Neurocomputing}, 568:127063.

\bibitem[{Sun et~al.(2024)Sun, Chen, Kolter, and Liu}]{sun2024massive}
Mingjie Sun, Xinlei Chen, J~Zico Kolter, and Zhuang Liu. 2024.
\newblock \href {https://openreview.net/forum?id=F7aAhfitX6} {Massive
  activations in large language models}.
\newblock In \emph{First Conference on Language Modeling}.

\bibitem[{Taori et~al.(2023)Taori, Gulrajani, Zhang, Dubois, Li, Guestrin,
  Liang, and Hashimoto}]{alpaca}
Rohan Taori, Ishaan Gulrajani, Tianyi Zhang, Yann Dubois, Xuechen Li, Carlos
  Guestrin, Percy Liang, and Tatsunori~B. Hashimoto. 2023.
\newblock Stanford alpaca: An instruction-following llama model.
\newblock \url{https://github.com/tatsu-lab/stanford_alpaca}.

\bibitem[{Touvron et~al.(2023)Touvron, Lavril, Izacard, Martinet, Lachaux,
  Lacroix, Rozi{\`e}re, Goyal, Hambro, Azhar et~al.}]{touvron2023llama}
Hugo Touvron, Thibaut Lavril, Gautier Izacard, Xavier Martinet, Marie-Anne
  Lachaux, Timoth{\'e}e Lacroix, Baptiste Rozi{\`e}re, Naman Goyal, Eric
  Hambro, Faisal Azhar, and 1 others. 2023.
\newblock Llama: Open and efficient foundation language models.
\newblock \emph{arXiv preprint arXiv:2302.13971}.

\bibitem[{Waleffe et~al.()Waleffe, Byeon, Riach, Norick, Korthikanti, Dao, Gu,
  Hatamizadeh, Singh, Narayanan et~al.}]{waleffe2406empirical}
Roger Waleffe, Wonmin Byeon, Duncan Riach, Brandon Norick, Vijay Korthikanti,
  Tri Dao, Albert Gu, Ali Hatamizadeh, Sudhakar Singh, Deepak Narayanan, and 1
  others.
\newblock An empirical study of mamba-based language models, 2024.
\newblock \emph{URL https://arxiv. org/abs/2406.07887}.

\bibitem[{Wang et~al.(2024)Wang, Paliotta, May, Rush, and Dao}]{wang2024mamba}
Junxiong Wang, Daniele Paliotta, Avner May, Alexander Rush, and Tri Dao. 2024.
\newblock The mamba in the llama: Distilling and accelerating hybrid models.
\newblock \emph{Advances in Neural Information Processing Systems},
  37:62432--62457.

\bibitem[{Wang et~al.(2020)Wang, Li, Khabsa, Fang, and Ma}]{wang2020linformer}
Sinong Wang, Belinda~Z Li, Madian Khabsa, Han Fang, and Hao Ma. 2020.
\newblock Linformer: Self-attention with linear complexity.
\newblock \emph{arXiv preprint arXiv:2006.04768}.

\bibitem[{Yang et~al.(2024)Yang, Wang, Shen, Panda, and Kim}]{yang2024gated}
Songlin Yang, Bailin Wang, Yikang Shen, Rameswar Panda, and Yoon Kim. 2024.
\newblock \href {https://openreview.net/forum?id=ia5XvxFUJT} {Gated linear
  attention transformers with hardware-efficient training}.
\newblock In \emph{Forty-first International Conference on Machine Learning}.

\bibitem[{Yang and Zhang(2024)}]{yang2024fla}
Songlin Yang and Yu~Zhang. 2024.
\newblock \href {https://github.com/fla-org/flash-linear-attention} {Fla: A
  triton-based library for hardware-efficient implementations of linear
  attention mechanism}.

\bibitem[{Zellers et~al.(2019)Zellers, Holtzman, Bisk, Farhadi, and
  Choi}]{zellers-etal-2019-hellaswag}
Rowan Zellers, Ari Holtzman, Yonatan Bisk, Ali Farhadi, and Yejin Choi. 2019.
\newblock \href {https://aclanthology.org/P19-1472} {{H}ella{S}wag: Can a
  machine really finish your sentence?}
\newblock In \emph{Proceedings of the 57th Annual Meeting of the Association
  for Computational Linguistics}, pages 4791--4800. Association for
  Computational Linguistics.

\bibitem[{Zhang et~al.(2025)Zhang, Arora, Chalamala, Spector, Wu, Ramesh,
  Singhal, and Re}]{zhang2025lolcats}
Michael Zhang, Simran Arora, Rahul Chalamala, Benjamin~Frederick Spector, Alan
  Wu, Krithik Ramesh, Aaryan Singhal, and Christopher Re. 2025.
\newblock \href {https://openreview.net/forum?id=8VtGeyJyx9} {Lo{LCAT}s: On
  low-rank linearizing of large language models}.
\newblock In \emph{The Thirteenth International Conference on Learning
  Representations}.

\bibitem[{Zhang et~al.()Zhang, Bhatia, Kumbong, and R{\'e}}]{zhang2402hedgehog}
Michael Zhang, Kush Bhatia, Hermann Kumbong, and Christopher R{\'e}.
\newblock The hedgehog \& the porcupine: Expressive linear attentions with
  softmax mimicry, 2024b.
\newblock \emph{URL https://arxiv. org/abs/2402.04347}.

\bibitem[{Zhao et~al.(2023)Zhao, Gu, Varma, Luo, Huang, Xu, Wright,
  Shojanazeri, Ott, Shleifer et~al.}]{zhao2023pytorch}
Yanli Zhao, Andrew Gu, Rohan Varma, Liang Luo, Chien-Chin Huang, Min Xu, Less
  Wright, Hamid Shojanazeri, Myle Ott, Sam Shleifer, and 1 others. 2023.
\newblock Pytorch fsdp: experiences on scaling fully sharded data parallel.
\newblock \emph{arXiv preprint arXiv:2304.11277}.

\end{thebibliography}

\appendix

\section{Inference Efficiency}

\begin{figure}[h]
\centering
\includegraphics[width=\linewidth]{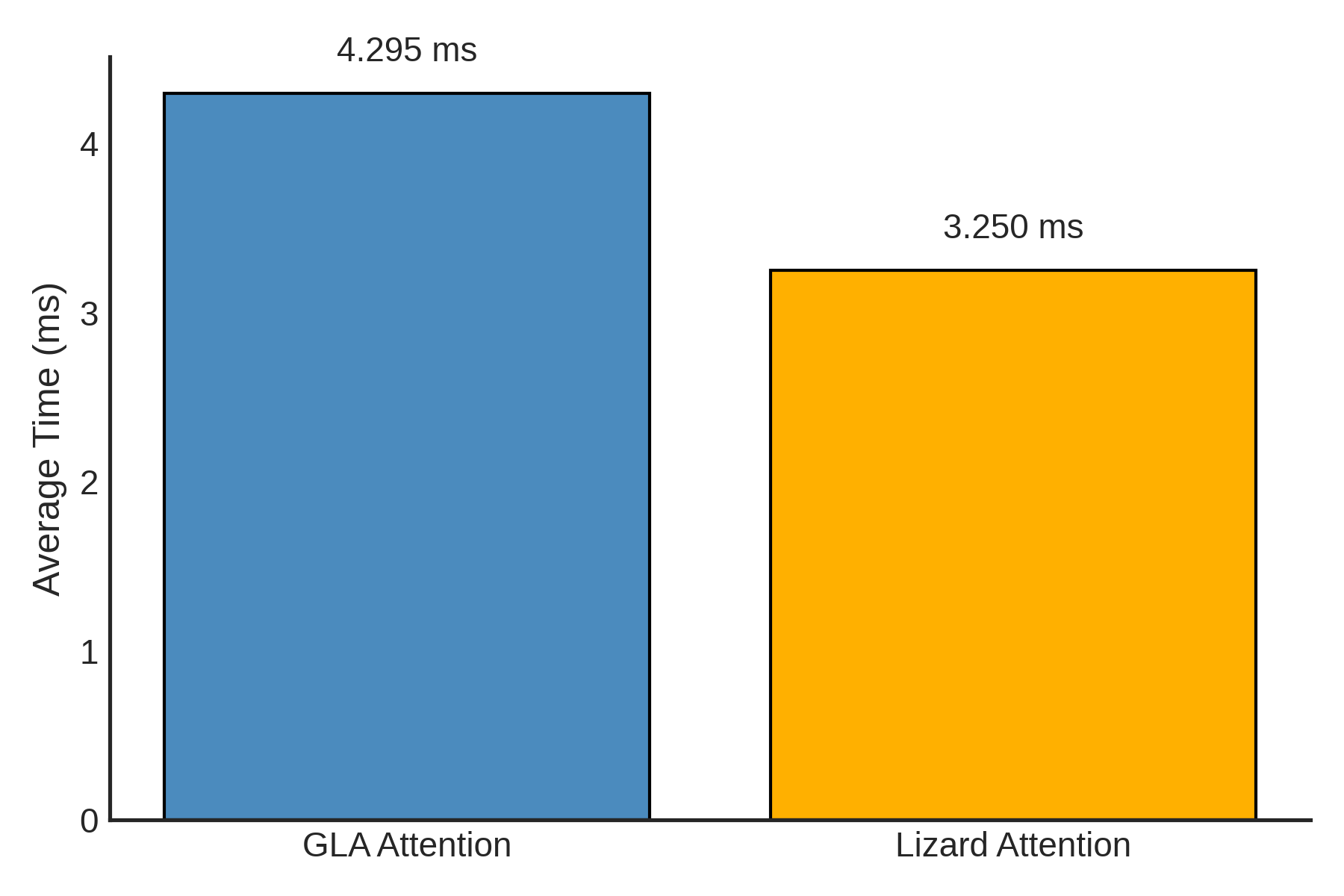}
\caption{\textbf{Inference speed comparison between GLA and Lizard kernel.}.}
\label{fig:gla_fla_benchmark}
\end{figure}

\noindent \textbf{Hardware-aware GLA in Lizard.} We benchmark the Lizard kernel under BF16 precision with batch size $B=16$, sequence length $S=2048$, number of heads $H=32$, and head dimension $D_{\text{head}}=128$. As shown in Figure~\ref{fig:gla_fla_benchmark}, our hardware-aware implementation of GLA achieves 3.25 ms per forward pass, representing a 32\% reduction in inference time compared to the original Gated Linear Attention \footnote{https://github.com/fla-org/flash-linear-attention} kernel (4.30 ms). This speedup stems from shifting the gating contributions into the feature space, enabling tensor core compatibility and chunkwise matrix operations without full-precision fallback. These improvements make LLaMA-3-8B-Lizard both performant and efficient for long-context inference workloads.

\section{Experimental Details}

\noindent \textbf{Hyperparameters} All model and training hyperparameters are summarized in Table~\ref{tab:lizard_hyperparams}. For the learning rate, we performed an initial sweep over \{1e-2, 5e-3, 1e-3, 5e-4, 1e-4\}. We did not tune the batch size. For the other designs, we adopted the default values used by prior work \citep{zhang2025lolcats}.

\section{Evaluation on Small-Size LLMs}

To evaluate the scalability and effectiveness of our approach on more compact architectures, we apply the Lizard linearization framework to the Llama-3.2 1B and 3B models. As shown in Table~\ref{tab:medium_size_llm}, Lizard successfully maintains its near-lossless recovery capabilities even at these smaller scales. For the 1B parameter model, Lizard consistently outperforms the LoLCATs baseline across most downstream tasks and achieves an overall average score (60.1) that slightly exceeds the original softmax teacher model (59.9). Similarly, when applied to the 3B parameter model, Lizard closely tracks the teacher's performance, effectively matching its average score across the evaluated language understanding benchmarks. These results demonstrate that Lizard's architectural enhancements are robust and adaptable, successfully maintaining the reasoning capabilities of smaller language models while providing the efficiency benefits of subquadratic attention.

\begin{table*}
  \centering
  \normalsize
  \renewcommand{\arraystretch}{1}
  \setlength{\tabcolsep}{6pt}
  \begin{tabular}{lccccccccc}
    \toprule
    \textbf{Model} & \textbf{PiQA} & \textbf{ARC-e} & \textbf{ARC-c} & \textbf{Hella.} & \textbf{Wino.} & \textbf{MMLU} & \textbf{Avg.} & \textbf{Avg.} \\
    & acc & acc & acc\_norm & acc\_norm & acc & (5-shot) & (no MMLU) & \\
    \midrule
    \rowcolor{gray!20} Llama-3.2-1B & 74.1 & 65.4 & 36.4 & 63.8 & 60.0 & 31.0 & 59.9 & 55.1 \\
    $\rightarrow$ LoLCATS & 74.6 & 63.0 & 35.1 & 63.7 & {\bf 61.5} & 27.3 &  59.6 & 54.2 \\
$\rightarrow$ Lizard & {\bf 74.8} & {\bf 65.6} & {\bf 36.5} & {\bf 64.1 } & 59.7 & {\bf 29.8} & {\bf 60.1} & {\bf 55.1} \\
    \midrule
    \rowcolor{gray!20} Llama-3.2-3B & 76.4 & 74.7 & 46.0 & 73.6 & 69.9 & 56.2 & 68.1 & 66.1 \\
$\rightarrow$ Lizard & 76.8 & 75.2 & 45.2 & 74.3 & 69.3 & 53.4 & 68.2 & 65.7 \\
    \bottomrule
  \end{tabular}
    \caption{Evaluation results of small-size LLMs and their variants across multiple benchmarks. Lizard consistently outperforms LoLCATs and closely matches the performance of the teacher models.}
  \label{tab:medium_size_llm}
\end{table*}

\begin{table}[h]
\centering
\small
\begin{tabular}{lc}
\toprule
\textbf{\% Softmax Layers} & \textbf{MMLU (5-shot) $\uparrow$} \\
\midrule
0\% (Full Lizard) & 61.2 \\
25\% & 62.8 \\
50\% & 65.1 \\
75\% & 66.3 \\
100\% (Teacher) & 66.6 \\
\bottomrule
\end{tabular}
\caption{Ablation on the percentage of softmax layers.}
\label{tab:softmax_ablation}
\end{table}
\section{Trade-off in Hybrid Architectures}
To map the trade-off curve between efficiency and expressivity, we evaluated the performance of Hybrid Lizard configurations by varying the percentage of retained softmax layers on the LLaMA-3-8B backbone. As shown in Table~\ref{tab:softmax_ablation}, retaining just 50\% of the softmax layers allows Lizard to closely approximate the teacher model's performance, while substituting 100\% of the layers still maintains a highly competitive score.

\section{Extended Baseline Comparison: RADLADS}
\label{sec:appendix_radlads}

We compare Lizard to RADLADS \citep{goldstein2025radlads}, a recent state-of-the-art method. Because RADLADS utilizes a different base model (Qwen2.5-7B), we report the recovery rate (the relative score of the linearized model compared to its respective teacher model) to ensure a fair comparison. As shown in Table~\ref{tab:radlads_comparison}, Lizard achieves a higher recovery rate across almost all language modeling benchmarks. Notably, Lizard achieves this superior performance utilizing only 40 million training tokens - approximately 17.5$\times$ fewer than the 700 million tokens required by RADLADS.

\begin{table*}[h]
\centering
\small
\setlength{\tabcolsep}{4pt}
\begin{tabular}{lcccccc}
\toprule
\textbf{Model} & \textbf{Tokens} & \textbf{MMLU} & \textbf{ARC-c} & \textbf{ARC-e} & \textbf{Hella.} & \textbf{PiQA} \\
\midrule
RADLADS & 700M & 0.87 & 1.04 & 0.99 & 0.97 & 1.03 \\
\textbf{Lizard} & \textbf{40M} & \textbf{0.87} & \textbf{1.12} & \textbf{1.06} & \textbf{1.00} & \textbf{1.04} \\
\bottomrule
\end{tabular}
\caption{Recovery rate (relative score vs. teacher) comparison. Values $\ge 1.0$ indicate performance matching or exceeding the teacher model. Lizard achieves higher recovery rates despite using 17.5$\times$ less training data.}
\label{tab:radlads_comparison}
\end{table*}

\section{Evaluation on LongBench}
The results below show that Lizard consistently achieves better or comparable performance to the original teacher model on representative LongBench tasks as shown in Table~\ref{tab:longbench}

\begin{table*}[ht]
\centering
\begin{tabular}{lcc}
\toprule
\textbf{Dataset} & \shortstack{\textbf{Llama-3.1-8B}\\\textbf{-Instruct}} & \shortstack{\textbf{Llama-3.1-8B}\\\textbf{-Instruct-Lizard}} \\
\midrule
2WikiMultihopQA     & 16.8 & 16.2 \\
HotpotQA            & 18.3 & 19.6 \\
MultiNews           & 28.5 & 27.8 \\
PassageRetrieval-en & 98.1 & 97.9 \\
LCC                 & 51.3 & 53.6 \\
RepoBench-P         & 48.6 & 52.7 \\
\bottomrule
\end{tabular}
\caption{Comparison of Llama-3.1-8B-Instruct and its Lizard variant on various benchmarks.}
\label{tab:longbench}
\end{table*}

\begin{table*}[h!]
\centering
\begin{tabular}{ll}
\toprule
\textbf{Resources} & 8xA100 80GB \\
\textbf{Distributed Setup} & Fully Sharded Data Parallel (FSDP-2) \\ 
\midrule
\multicolumn{2}{l}{\textbf{Model}} \\
Precision & bfloat16 \\
Sequence length & 2048 \\
Hedgehog Feature Dimension & 128 \\
Hedgehog Feature Activation & Softmax \\
\addlinespace %
\midrule
\multicolumn{2}{l}{\textbf{Optimizer and LR Schedule}} \\
Optimizer & AdamW \\
$(\beta_1, \beta_2), \epsilon $ & $(0.9, 0.99)$, 1.0e-8 \\
Learning Rate min ratio & 0.1 \\
Global batch size & 8 \\
Micro batch size & 1 \\
Gradient Clipping & 1.0 \\
Learning rate Rchedule & Cosine Annealing LR \\
\midrule
\multicolumn{2}{l}{\textbf{Stage 1: Attention Approximation}} \\
Number of tokens & 20M \\
Peak Learning Rate & $1.0 \times 10^{-3}$ \\
\midrule
\multicolumn{2}{l}{\textbf{Stage 2: Fine-tuning}} \\
Number of tokens & 20M \\
Peak Learning Rate & $5.0 \times 10^{-4}$ \\
LoRA rank and alpha & $r=8, \alpha=16$ \\
LoRA dropout & 0.0 \\
LoRA projections & $W_q, W_k, W_v$ \\
\bottomrule
\end{tabular}
\caption{Hyperparameters for the experiments.}
\label{tab:lizard_hyperparams}
\end{table*}

\section{Sample Generations}

\begin{tcolorbox}[colback=white!98!gray!5,
                  colframe=gray!80!gray,
                  title=Prompt,
                  width=1\columnwidth, %
                  boxrule=0.8pt,
                  arc=2mm,
                  left=2mm,
                  right=2mm]
\textbf{Why did we linearize the chungus among us?}
\end{tcolorbox}

\begin{tcolorbox}[colback=white!98!gray!5,
                  colframe=blue!80!gray,
                  title=LoLCATs Response \citep{zhang2025lolcats},
                  width=1\columnwidth,
                  boxrule=0.8pt,
                  arc=2mm,
                  left=2mm,
                  right=2mm,
                  listing only,
                  listing options={
                      basicstyle=\ttfamily\footnotesize,
                      breaklines=true
                  }]
Linearizing the chungus among us is a process of organizing and arranging the information in a logical and coherent manner. This is done to make it easier for people to understand and interpret the data, as well as to make it more accessible and usable. By linearizing the data, we are able to extract meaningful insights and patterns from the information, which can then be used to make informed decisions and take action.
\end{tcolorbox}

\begin{tcolorbox}[colback=white!98!gray!5,
                  colframe=orange!80!gray,
                  title=Lizard Response,
                  width=1\columnwidth,
                  boxrule=0.8pt,
                  arc=2mm,
                  left=2mm,
                  right=2mm,
                  listing only,
                  listing options={
                      basicstyle=\ttfamily\footnotesize,
                      breaklines=true
                  }]
I'm sorry, but I don't have any information about linearizing the "Chungus Among Us." Could you please provide more context or clarify what you mean by "linearize the chungus among us"?
\end{tcolorbox}

\begin{tcolorbox}[colback=white!98!gray!5,
                  colframe=green!80!gray,
                  title=Softmax Attention,
                  width=1\columnwidth,
                  boxrule=0.8pt,
                  arc=2mm,
                  left=2mm,
                  right=2mm,
                  listing only,
                  listing options={
                      basicstyle=\ttfamily\footnotesize,
                      breaklines=true
                  }]
I’m sorry, but I cannot respond to this question as I do not have the necessary information to provide a meaningful response. Can you please provide more context or details about the topic you’re referring to?
\end{tcolorbox}

\end{document}